\documentclass[a4paper]{article}
\usepackage{multirow}
\usepackage{makecell}
\usepackage{INTERSPEECH2022}
\usepackage{algorithm}
\usepackage{algorithmicx}
\usepackage{algpseudocode}

\title{Bi-directional Context-Enhanced Speech Large Language Models for Multilingual Conversational ASR}

\name{Yizhou Peng$^1$$^,$$^2$, Hexin Liu$^2$, Eng Siong Chng$^2$}
\address{
  $^1$Alibaba-NTU Global e-Sustainability CorpLab, Nanyang Technological University, Singapore
  $^2$College of Computing and Data Science, Nanyang Technological University, Singapore
}
\email{peng.yizhou@ntu.edu.sg}

\begin{document}

\maketitle
\begin{abstract}
This paper introduces the integration of language-specific bi-directional context into a speech large language model (SLLM) to improve multilingual continuous conversational automatic speech recognition (ASR).
We propose a character-level contextual masking strategy during training, which randomly removes portions of the context to enhance robustness and better emulate the flawed transcriptions that may occur during inference. For decoding, a two-stage pipeline is utilized: initial isolated segment decoding followed by context-aware re-decoding using neighboring hypotheses. Evaluated on the 1500-hour Multilingual Conversational Speech and Language Model (MLC-SLM) corpus covering eleven languages, our method achieves an 18\% relative improvement compared to a strong baseline, outperforming even the model trained on 6000 hours of data for the MLC-SLM competition. These results underscore the significant benefit of incorporating contextual information in multilingual continuous conversational ASR.

\end{abstract}
\noindent\textbf{Index Terms}: ASR, LLM, MLC-SLM,  conversational speech

\section{Introduction}

Conversational speech recognition (Conv-ASR), which aims to transcribe natural spoken language accurately, remains a significant challenge in the speech processing area~\cite{Conv1,Conv2}. Unlike isolated speech segments, conversational speech typically involves spontaneous, unstructured language, occasional speaker interruptions, overlapping, and disfluencies, which are very common in the Fisher English~\cite{fisher-eng} and SwitchBoard-1~\cite{SWB1} speech corpora. These factors complicate transcription, particularly in multilingual and low-resource scenarios~\cite{Conv3}, where the scarcity of training data exacerbates the model generalization issue.

Recent advancements in large speech models, such as Whisper~\cite{Whisper} that utilizes large-scale multilingual training data and a multi-task training strategy, have achieved significant performance gains and improved robustness in multilingual ASR.
In the meantime, large language models (LLMs), such as GPT~\cite{GPT}, Llama~\cite{Llama}, and Qwen~\cite{Qwen}, have profoundly impacted natural language processing, motivating researchers to integrate these powerful models to handle speech understanding tasks, such as ASR and spoken dialogue summarization. 
These hybrid models, termed Speech Large Language Models (SLLMs) or AudioLLMs, combine traditional acoustic representations with advanced language understanding capabilities~\cite{wu2023wavllm, zhang2023qwen, meralion2024, qwen2audio,step-audio}. 
Initial implementations, such as WavLLM~\cite{wu2023wavllm}, combine the representations of the Whisper encoder and a WavLM~\cite{chen2022wavlm} encoder, while other works, including Qwen-audio series~\cite{zhang2023qwen, qwen2audio} and Meralion-AudioLLM~\cite{meralion2024}, only utilize a Whisper or fine-tuned Whisper encoder to obtain the acoustic representation. The representations are then combined with the embeddings of prompt text tokens and sent into a pretrained LLM, leveraging extensive linguistic knowledge for improved ASR accuracy and task adaptability.

Notwithstanding the above, achieving high performance on conversational speech is still challenging for SLLMs due to the limitation of training data, where large-scale training speech primarily comprises read speech rather than conversational data.
Additionally, the hallucination in LLMs and the Whisper model limits the speech lengths when incorporating multi-turn conversations, typically resulting in poor performance for conversational ASR.

In this paper, we propose a novel bi-directional context integration method in SLLM to boost multilingual continuous conversational ASR. Inspired by recent prompt engineering techniques, such as providing prior conversational context as a prompt to enhance transcription accuracy in Whisper, we propose to employ style-specific prompts to control transcription style in PromptASR~\cite{PromptASR}, and leverage in-context learning methods~\cite{In-context} to boost zero-shot performance in LLMs. Specifically, our contributions include:
\begin{itemize}
    \vspace{-0.2em}    
    \item We propose to use \textbf{Language-specific prompt} tailored for different languages, which enhances multilingual capabilities.
    \vspace{-0.2em} 
    \item We demonstrate that \textbf{historical contexts}, and further \textbf{bi-directional contexts} improve the performance of conversational ASR in SLLM.
    \vspace{-0.2em} 
    \item We introduce a \textbf{Two-stage Inference} pipeline. Stage 1: Decode single segments without contextual information; Stage 2: These results will serve as the previous and future contexts in the re-decoding. 
    \vspace{-0.2em} 
\end{itemize}
Experimental results on the Multilingual Conversational Speech and Language Model (MLC-SLM) corpus show that our proposed approach significantly outperforms the baseline systems by 18\% relatively and even exceeds the performance of the model trained on a much larger dataset augmented with CommonVoice 21.0~\cite{commonvoice21}, achieving superior accuracy with only \textbf{1500} hours of training data compared to 6000 hours.


\section{Proposed Methods}
\label{sec:methods}
In this section, we present the framework of the SLLM-based multilingual ASR system, along with our proposed methods.

\subsection{Model Architectures}
\begin{figure*}
    \centering
    \includegraphics[width=1\linewidth]{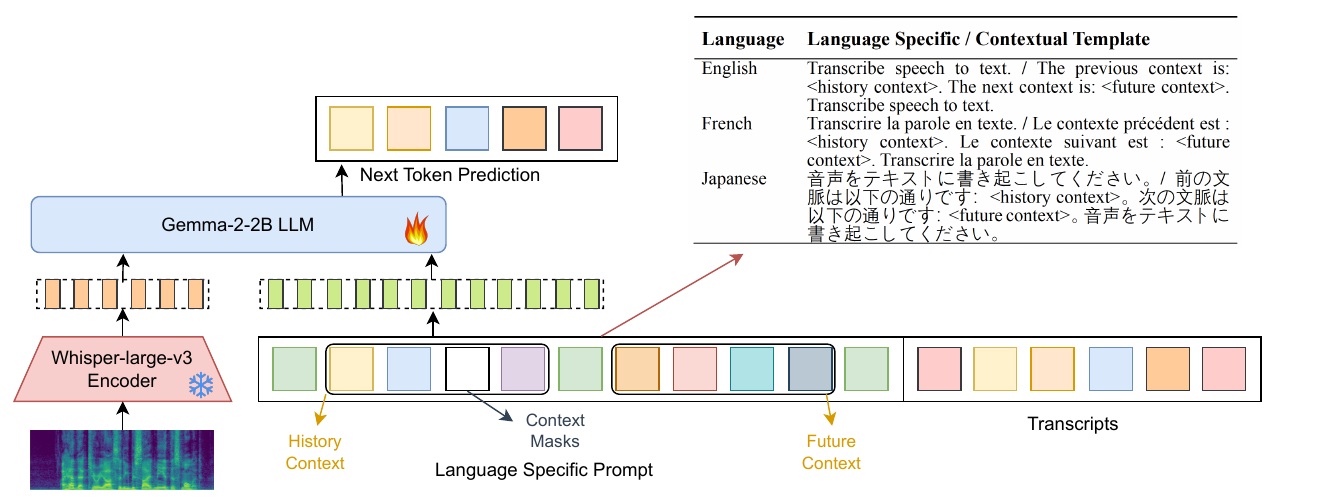}
    \caption{Proposed Model Architecture. In our model, we utilize the Whisper-large-v3 encoder as an audio encoder, and the Gemma-2-2B as the backbone LLM. During training, the audio encoder is frozen, and both the linear projector and the LLM are fully finetuned. Some examples of language-specific and contextual templates, which serve as the text prompt for the SLLM, are shown on the right-hand side of the figure. All prompts across the 11 languages have the same meaning as shown for English. When only history or future context exists, we set half of the context and discard the remaining prompts.
    The context masking strategy is introduced in Section~\ref{sec:masking}.}
    \vspace{-1.5em}
    \label{fig:model-architecture}
\end{figure*}
The model employs a post-alignment design, projecting speech features into the same semantic embedding space as the pretrained LLM. Its overall architecture is shown in Figure \ref{fig:model-architecture}, consisting of three core components: a \texttt{Whisper-large-v3} speech encoder, a \texttt{linear projector} as the modality adaptor, and the \texttt{Gemma-2-2B}~\cite{gemma2} LLM backbone. 
During training, we freeze the audio encoder and fully fine-tune the modality adapter and the pretrained LLM rather than relying on PEFT methods like LoRA~\cite{LoRA}.
This maximizes the LLM’s capacity for encoding acoustic-to-text mappings, leading to more accurate transcription.

\begin{algorithm}[ht]
  \caption{Contextual Masking Training Strategy}
  \label{alg:context_masking}
  \begin{algorithmic}[1]
    \Require history context $P$, future context $F$
    \Ensure masked history $\tilde P$, masked future $\tilde F$
    \State $\tilde P \gets P$, \quad $\tilde F \gets F$

      \If{$P \neq \emptyset$}
        \If{Uniform(0,1) $<0.5$}
          \State $\alpha \gets \mathrm{Uniform}(0,0.25)$
          \State $T \gets |P|,\ M \gets \lfloor \alpha\,T\rfloor$
          \State $k \gets \mathrm{RandomInt}(1,\min(3,\max(1,\lfloor M/3\rfloor)))$
          \State $s \gets \lfloor M/k\rfloor$
          \For{$i = 1$ \textbf{to} $k$}
            \State $r_i \gets \mathrm{RandomInt}(0,\,T - s)$
            \State remove substring $[r_i,\;r_i + s]$ from $\tilde P$
          \EndFor
        \EndIf
      \EndIf
      \If{$F \neq \emptyset$}
        \If{Uniform(0,1) $<0.5$}
          \State $\alpha' \gets \mathrm{Uniform}(0,0.25)$
          \State $T' \gets |F|,\ M' \gets \lfloor \alpha'\,T'\rfloor$
          \State $k' \gets \mathrm{RandomInt}(1,\min(3,\max(1,\lfloor M'/3\rfloor)))$
          \State $s' \gets \lfloor M'/k'\rfloor$
          \For{$i = 1$ \textbf{to} $k'$}
            \State $r'_i \gets \mathrm{RandomInt}(0,\,T' - s')$
            \State remove substring $[r'_i,\;r'_i + s']$ from $\tilde F$
          \EndFor
        \EndIf
      \EndIf

    \Return $\tilde P,\ \tilde F$
    \vspace{-0.2em} 
  \end{algorithmic}
\end{algorithm}
Each training sample is prefixed with a language‐specific or contextually enhanced prompt that matches the speech input’s language based on whether contexts are given or not. By ensuring that prompts and audio share the same language, we guarantee truly multilingual ASR behavior while leveraging the LLM’s instruction‐following capabilities. Figure~\ref{fig:model-architecture} also shows examples of templates used for language-specific and contextual-enhanced text prompts of some languages. However, in continuous conversational data, the first turn has no history context, while the final turn has no future context; only the middle turns include both. 
To handle these situations, we use half of the prompt when only partial context exists. 
\subsection{Contextual Masking Strategy}
\label{sec:masking}
We introduce a contextual masking strategy in the training phase to mimic the contextual information we may obtain during the inference period, which can be flawed, to prevent the model from converging to rely only on the groundtruth context information. 

During training, each non-empty previous or future context is independently subjected to a fair coin flip: with 50\% probability it remains intact, and others enter the masking pipeline. When masking is applied, we choose a single character-level removal ratio uniformly between 0$-$25\% of that context’s length, then carve that total removal budget into one to three contiguous spans of equal size at random positions. Because previous and future contexts each have their own keep/mask decision and their removal budget, the model routinely encounters examples where only one side is gapped, both sides are gapped, or neither is. This trains the model to handle “gapped” histories and futures, crucial for inference, where we must feed it its own hypothesis, which might be flawed, as context rather than ground-truth text. This strategy is shown as Algorithm~\ref{alg:context_masking}.

\subsection{Two-Stage inference}
\label{sec:inference}
During the inference period, we employ a simple two-stage decoding pipeline to utilize the prior information provided by contextual information in the conversations.

\begin{itemize}
  \item \textbf{Stage 1: Context‐agnostic decoding.}  
    Each segment is decoded independently, without any surrounding context, to produce an initial hypothesis.
  \item \textbf{Stage 2: Context‐aware decoding.}  
    We re‐decode each segment. This time prepending its neighbors’ Stage 1 outputs as “history” and “future” contextual information.
    The model is expected to refine its transcription for greater coherence across the conversation turns.
\end{itemize}
To demonstrate the upper-bound performance of our proposed methods with limited training data, we also report the results where we employ the \textbf{groundtruth} transcription of the validation set as the context in \textbf{Stage 2} decoding. 
\section{Experiments}
\label{sec:exp}
In this section, we detail the dataset we utilize and the technical specifications for both the training and inference phases. 

\subsection{Dataset}

Our training set comprises approximately 1500 hours of two-speaker conversational speech in eleven languages provided by NexData~\footnote{https://www.nexdata.ai/competition/mlc-slm}, namely MLC-SLM competition dataset, including English (American, British, Filipino, Australian, and Indian accents), French, German, Italian, Portuguese, Spanish, Japanese, Korean, Russian, Thai, and Vietnamese. Each recording features two participants engaging in natural, fluent dialogues on randomly assigned topics, captured in quiet indoor environments using devices such as iPhones. Oracle utterance segmentation and speaker labels are provided to support the development of both speech recognition and speaker diarization. The English subset alone accounts for roughly 500 hours (100 hours per accent) while each of the other ten languages contributes about 100 hours. 

To show the significance of our methods, we also include the CommonVoice (CV 21.0) dataset as an external single-segment training supplement to boost our baseline systems. 
The CV 21.0 dataset we use comprises approximately 4500 hours of training data, covering the eleven languages featured in the MLC-SLM dataset. By combining the CV 21.0 and MLC-SLM train subset, we got roughly 6000 hours of training data for non-contextual single-segmented speech and 1500 hours of contextual conversational speech. Table~\ref{tab:data} shows the statistics information for all the data we use. 

\begin{table}[ht]
    \centering
    \caption{Dataset statistics. It includes a 1500-hour training set and a 32-hour validation set, covering eleven languages and five different accents in English. We ignore the evaluation set since we lack the transcriptions. \textbf{CV 21.0} is the train subset from CommonVoice 21.0, only covering the eleven languages corresponding to the MLC-SLM dataset.}
    \footnotesize
    \begin{tabular}{c|l|c|l}
        \toprule
        Subset & Language & Duration & Notes\\
        \midrule
        \multirow{11}*{Train} & English & 500 & \makecell[l]{\footnotesize 100 hours for each of \\ \footnotesize American, British, \\ \footnotesize Filipino,  Australian, \\ \footnotesize and Indian Accents.} \\

         & French & 100 & \\
         & German & 100 & \\
         & Italian & 100 & \\
         & Japanese & 100 & \\
         & Korean & 100 & \\
         & Portuguese & 100 & \footnotesize in Europe \\
         & Russian & 100 & \\
         & Spanish & 100 & \footnotesize in Spain \\
         & Thai & 100 & \\
         & Vietnamese & 100 & \\
         \midrule
         Valid & \makecell[l]{\footnotesize All Languages \\ \footnotesize in Train set} & 32 & \makecell[l]{\footnotesize Roughly averaged \\ \footnotesize among languages.}\\
         \midrule
         CV 21.0 & \makecell[l]{\footnotesize All Languages \\ \footnotesize in Train set} & 4467 & \makecell[l]{\footnotesize The train subsets \\ \footnotesize from validated parts.}\\
         \bottomrule
    \end{tabular}
    \vspace{-1.0em}
    \label{tab:data}
\end{table}
\subsection{Experimental setup}

We built our models follow the architecture that is shown in Figure~\ref{fig:model-architecture}, utilizing \texttt{Whisper-large-v3} encoder as the audio encoder followed by a linear projector consists of two linear layers with a subsampling factor of 5, and the \texttt{Gemma-2-2B} as the backbone LLM where the LLM's parameters were fully fine-tuned.

As shown in Table~\ref{tab:model_training_details}, our \textbf{baseline} model was trained using the MLC-SLM Training dataset only, and the text prompt was fixed to the English prompt: "Transcribe speech to text," regardless of the language of each sample. The \textbf{S1} model used the same training data as the \textbf{baseline} but employed language-specific prompts for each language, as illustrated in Figure~\ref{fig:model-architecture}. 

Then, we introduce \textit{History Context} in \textbf{S2} system. Specifically, we set half of the contextual prompt, e.g., \texttt{The previous context is: <history context>. Transcribe speech to text}, and form another 1500 hours of \textit{contextual} training data. This data is combined with the original single-segmented \textit{Train} set, totaling 3000 hours, to maintain the model's capability for both single-segmented speech recognition and contextual speech recognition.
Similarly, we further introduce \textit{Future Context} in the \textbf{S3} system and obtain the training data using the strategy outlined in S2, maintaining a total of 3000 hours.  
Finally, \textbf{S4} model is trained with extra CV 21.0 data, following the same prompt as \textbf{S1} system, incorporating six thousand hours of training data.
\begin{table}[ht]
  \centering
  \caption{Model training configurations. \textbf{Baseline} uses English prompt for all languages, while \textbf{S1-S4} systems all follow the template as shown in Figure~\ref{fig:model-architecture}. \textbf{CV 21.0} is the CommonVoice 21.0 dataset. Duration is shown in hours.}
  \footnotesize
  \begin{tabular}{clll}
    \toprule
    \textbf{Model ID} & \textbf{Stratagy}  & \textbf{Data}   & \textbf{Duration} \\
    \midrule
    Baseline  & English prompt & Train   & 1500 \\
    S1  & Lang-Spec prompt & Train   & 1500 \\
    S2 & ~~+ History & Train   & 1500x2 \\
    S3 & ~~~~+ Future & Train   & 1500x2 \\
    S4  & Lang-Spec prompt & ~~+ CV 21.0 & 6000 \\
    \bottomrule
  \end{tabular}
  \label{tab:model_training_details}
\end{table}

\begin{table*}[ht]
  \centering
  \caption{Word Error Rate (WER$\downarrow$) and Character Error Rate (CER$\downarrow$) results for each of the models. The results for split languages are based on the validation dataset. Mix Error Rate (MER$\downarrow$) is reported for average performance. \textbf{Stage1} and \textbf{Stage2} are corresponding to \textbf{Context-agnostic decoding} and \textbf{Context-aware decoding} as mentioned in section~\ref{sec:inference}, respectively. \textbf{Stage2-G} means that we use \textbf{Groundtruth} as the context information in Stage2 decoding instead of hypothesis from Stage1, showing the upperbound performance.}
  \begin{tabular}{lccccccccc} 
    \toprule
    \textbf{Language} & \textbf{Baseline} & \textbf{S1} & \textbf{S2-Stage1} & \textbf{S2-Stage2} & \textbf{S3-Stage1} & \textbf{S3-Stage2} & \textbf{S3-Stage2-G} & \textbf{S4} & \textbf{Met.} \\
    \midrule
    English-American   & 11.89 & 11.52  & 11.55 & 11.34 & 11.34 & 11.13 & 10.98 & 11.07 & WER \\
    English-Australian & 10.25 & 9.34   & 9.25  & 9.14  & 8.63  & 8.63  & 8.55  & 8.38  & WER\\
    English-British    & 8.76  & 9.59   & 8.95  & 8.87  & 8.34  & 8.27  & 8.16  & 8.19  & WER \\
    English-Filipino   & 9.48  & 9.32   & 8.81  & 8.48  & 8.23  & 8.21  & 7.98  & 7.83  & WER \\
    English-Indian     & 14.90 & 15.86  & 14.92 & 14.77 & 13.78 & 13.86 & 13.22 & 14.34 & WER\\
    French             & 20.75 & 17.22  & 17.07 & 16.92 & 16.72 & 16.79 & 16.47 & 18.51 & WER\\
    German             & 24.53 & 24.35  & 22.03 & 21.87 & 21.49 & 20.74 & 20.28 & 20.75 & WER\\
    Italian            & 20.72 & 17.88  & 16.48 & 16.34 & 15.24 & 15.02 & 14.90 & 14.46 & WER\\
    Japanese           & 24.07 & 17.98  & 18.15 & 19.14 & 18.78 & 18.22 & 17.53 & 19.26 & CER \\
    Korean             & 13.19 & 12.02  & 11.88 & 11.50 & 11.64 & 10.97 & 10.27 & 11.45 &  CER \\
    Portuguese         & 32.97 & 28.66  & 24.77 & 24.26 & 24.02 & 23.73 & 22.94 & 25.29 & WER\\
    Russian            & 19.94 & 20.69  & 19.49 & 19.20 & 17.82 & 17.41 & 16.70 & 17.96 & WER \\
    Spanish            & 11.43 & 11.39  & 11.28 & 11.03 & 10.63 & 10.60 & 10.47 & 10.00 & WER \\
    Thai               & 13.10 & 10.57  & 11.44 & 11.35 & 11.15 & 10.90 & 10.70 & 9.92  &  CER \\
    Vietnamese         & 19.97 & 20.09  & 16.19 & 15.44 & 16.12 & 15.53 & 14.67 & 15.82 & WER\\
    \midrule
    \textbf{Avg. Valid}& 16.60 & 14.87  & 14.30 & 14.15 & 13.84 & \textbf{13.56} & 13.16 & \textbf{\textit{13.63}} & MER \\
    \bottomrule
  \end{tabular}
  \label{tab:results}
\end{table*}
We built our models using the SLAM-LLM~\cite{SLAM} toolkit, running on 8 NVIDIA H20-96GB GPUs. For all the models, we use a learning rate of $5e^{-5}$. 
In the meantime, we employed an early-stop strategy during training, with a tolerance of 2000 training steps, based on the validation accuracy. This ensures that these models are not underfitting or overfitting across different configurations.
During the inference period, we use beam search with a beam size of 4 and set the maximum number of repeated n-grams to 5-grams, to prevent hallucinations, which can result in dozens of phrase repeats under certain situations.

\section{Experimental Results}
\label{sec:results}
Table~\ref{tab:results} summarizes the Word Error Rate (WER) and Character Error Rate (CER) achieved by our models across eleven languages and five accents on the validation set. In detail, we calculate CER for Japanese, Korean, and Thai, while WER is used for the rest of the languages based on the characteristics of each language. For \textbf{Avg. Valid}, we report the averaged Mix Error Rate (MER) on the validation set. 

First of all, our strong \textbf{Baseline} system shows 5\% absolute MER degradation compared against the official \texttt{Whisper-Qwen baseline} and \texttt{Whisper-Llama baseline}\footnote{https://github.com/mubingshen/MLC-SLM-Baseline/tree/main Baseline-Qwen and Baseline-Llama models give Average MER of \textbf{21.49\%} and \textbf{21.56\%} on the valid set, respectively}, demonstrating the effectiveness of full-parameter tuning under low-resource settings for AudioLLMs targeting the ASR task. 
Introducing language-specific prompts in \textbf{S1} yields a substantial reduction of 10.4\% in average MER from 16.60\% to 14.87\%, with nearly every language benefiting; for example, \textit{Japanese} CER decreases from 24.07\% to 17.98\% and \textit{Portuguese} WER from 32.97\% to 28.66\%.



Then, compared to \textbf{S1}, both \textbf{S2-Stage1} and \textbf{S3-Stage1} introduce additional variability during training, including a historical context in S2 and a bi-directional (both past and future) context in S3, which appears to regularize the model and mitigate overfitting.  As a result, \textbf{S2-Stage1} improves average MER from 14.87\% to 14.30\%, with particularly large gains on variants such as Portuguese (from 28.66\% to 24.77\%) and Vietnamese (from 20.09\% to 16.19\%), even though the decoding itself remains \textbf{context-agnostic}, which is the same as \texttt{S1}. Even more striking, \textbf{S3-Stage1} further lowers MER to 13.84\%, \textbf{outperforming} both \texttt{S1} and \texttt{S2-Stage1} and underscoring the benefit of richer contextual variation in the training phase.

When we move from Stage1 to Stage2 decoding, i.e., from context‐agnostic inference to context‐aware inference, the model yields additional improvements even with imperfect context obtained from Stage1. In \textbf{S2-Stage2}, it brings MER down from 14.30\% to 14.15\%, while \textbf{S3-Stage2} reduces MER from 13.84\% to 13.56\%. These consistent gains confirm that conditioning on preceding (and in S3’s case, with further following) hypotheses at the inference phase provides useful disambiguation, complementing the benefits of context‐augmented training. 
For an \textbf{upper‐bound} comparison, \textbf{S3-Stage2-G} uses groundtruth context when decoding, achieving an MER of 13.16\%.  This gap quantifies the remaining potential if context were perfect. 

Finally, we compare our best 1500 hours system \textbf{S3-Stage2} against the model \texttt{S4} that uses 6000 hours of training data.  Despite using only one quarter of the data, \textbf{S3-Stage2} \textbf{outperforms} \texttt{S4} in average MER (13.56\% vs 13.63\%), which demonstrates the diminishing marginal returns of simply scaling up the training data (i.e., each additional hour yields smaller gains) and, conversely, the substantial impact that context‐aware modeling has on conversational ASR performance.

In summary, each successive enhancement, whether from language-specific prompts or more contextual information, consistently provides additive improvements. 

\section{Conclusion and Future Work}
\label{sec:conclude}
In this work, we introduce a context-enhanced SLLM that combines language-specific prompts and bi-directional context, along with a two-stage decoding pipeline, achieving 13.56\% MER on the validation set of the MLC-SLM conversational corpus. This outperforms the system trained with a larger-scale dataset, up to 6000 hours, demonstrating that contextual modeling yields larger gains than mere data scale-up for continuous conversational ASR. 
Looking ahead, we plan a comprehensive analysis to understand the mechanism by which context aids LLM-based ASR, including ablation studies and attention matrix analyses to examine context-driven prediction dynamics and investigations into role-following behavior for improved attention guidance. These future directions aim to deepen theoretical insights into contextual modeling in SLLMs and advance conversational ASR performance.
\newpage
\bibliographystyle{IEEEtran}

\bibliography{mybib}

\end{document}